\documentclass[11pt,a4paper]{article}
\usepackage[hyperref]{naaclhlt2018}
\usepackage{hyperref}
\usepackage{times}
\usepackage{latexsym}
\usepackage{url}
\usepackage[T1]{fontenc}
\usepackage{graphicx}
\usepackage{eurosym}
\usepackage{tabularx}
\graphicspath{ {images/} }
\usepackage{listings}
\usepackage{stfloats}
\usepackage{csquotes}
\usepackage{array}
\usepackage[super]{nth}
\usepackage[inline,shortlabels]{enumitem}
\usepackage{subfig}
\usepackage{amsmath}
\usepackage{multirow}
\usepackage{dirtytalk}
\usepackage{float}
\usepackage[scientific-notation=true]{siunitx}
\usepackage[textsize=tiny]{todonotes}
\usepackage{mdframed}
\usepackage{xprintlen}

\DeclareMathOperator*{\argmax}{arg\,max} 
\aclfinalcopy 
 
\title{NTUA-SLP at SemEval-2018 Task 3: Tracking Ironic Tweets using Ensembles of Word and Character Level Attentive RNNs}

\author{ 
	Christos Baziotis$^{1,3}$,  Nikos Athanasiou$^1$, Pinelopi Papalampidi$^1$, \\ 
	{\bf Athanasia Kolovou$^{1,2}$, Georgios Paraskevopoulos$^{1,4}$, Nikolaos Ellinas$^1$} \\
    {\bf Alexandros Potamianos$^{1,4}$}
    \\\\
	$^1$School of ECE, National Technical University of Athens, Athens, Greece \\
	$^2$ Department of Informatics, University of Athens, Athens, Greece \\
	$^3$ Department of Informatics, Athens University of Economics and Business, Athens, Greece \\
	$^4$ Behavioral Signal Technologies, Los Angeles, CA\\       
	{\tt cbaziotis@mail.ntua.gr, el12074@central.ntua.gr} \\
	{\tt el12003@central.ntua.gr, akolovou@di.uoa.gr} \\
	{\tt geopar@central.ntua.gr, nellinas@central.ntua.gr} \\
	{\tt potam@central.ntua.gr}
}

\date{2018}

\begin{document}

\maketitle


\begin{abstract}
In this paper we present two deep-learning systems that competed at SemEval-2018 Task 3 \enquote{Irony detection in English tweets}. 
We design and ensemble two independent models, based on recurrent neural networks (Bi-LSTM), which operate at the word and character level, in order to capture both the semantic and syntactic information in tweets. Our models are augmented with a self-attention mechanism, in order to identify the most informative words. 
The embedding layer of our word-level model is initialized with word2vec word embeddings, pretrained on a collection of 550 million English tweets.
We did not utilize any handcrafted features, lexicons or external datasets as prior information and our models are trained end-to-end using back propagation on constrained data. Furthermore, we provide visualizations of tweets with annotations for the salient tokens of the attention layer that can help to interpret the inner workings of the proposed models.
We ranked \nth{2} out of 42 teams in Subtask A and \nth{2} out of 31 teams in Subtask B. However, post-task-completion enhancements of our models achieve state-of-the-art results ranking \nth{1} for both subtasks.
\end{abstract}

\section{Introduction}
Irony is a form of figurative language, considered as \say{saying  the  opposite  of what  you  mean},  where  the opposition  of  literal  and  intended  meanings  is very  clear \cite{barbieri2014automatic,liebrecht2013perfect}. Traditional approaches in NLP \cite{tsur2010icwsm,barbieri2014automatic,karoui2015towards,farias2016irony} model irony based on pattern-based features, such as the contrast between high and low frequent words, the punctuation used by the author, the level of ambiguity of the words and the contrast between the sentiments. Also, \cite{joshi2016word} recently added word embeddings statistics to the feature space and further boosted the performance in irony detection. \par

\begin{figure}[!t]

  \begin{mdframed}
    \captionsetup{farskip=0pt} 
    \subfloat{\includegraphics[scale=0.95,page=33]{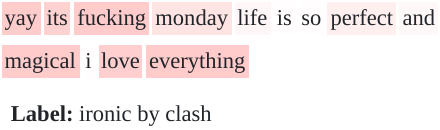}\label{fig:intro_att_1}}
    \bigbreak
    \subfloat{\includegraphics[scale=0.95,page=6]{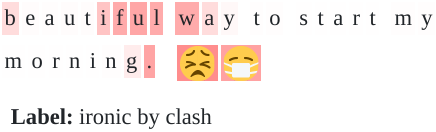}\label{fig:intro_att_2}}
  \end{mdframed}

	\caption{Attention heat-map visualization. 
    The color intensity of each word / character, corresponds to its weight (importance), as given by the self-attention mechanism (Section \ref{sec:self-att}).}
	\label{fig:intro-att}
\end{figure}

Modeling irony, especially in Twitter, is a challenging task, since in ironic comments literal meaning can be  misguiding; irony is expressed in ``secondary'' meaning and fine nuances that are hard to model explicitly in machine learning algorithms. Tracking irony in social media posses the additional challenge of dealing with special language, social media markers and abbreviations. Despite the accuracy achieved in this task by hand-crafted features, a laborious feature-engineering process and domain-specific knowledge are required; 
this type of prior knowledge must be continuously updated and investigated for each new domain. Moreover, the difficulty in parsing tweets \cite{gimpel2011} for feature extraction renders their precise semantic representation, which is key of determining their intended gist, much harder. \par
In recent years, the successful utilization of deep learning architectures in NLP led to alternative approaches for tracking irony in Twitter \cite{joshi2017automatic,ghosh2017magnets}. \cite{ghosh2016fracking} proposed a Convolutional Neural Network (CNN) followed by a Long Short Term Memory (LSTM) architecture, outperforming the state-of-the-art. \cite{dhingra2016tweet2vec} utilized deep learning for representing tweets as a sequence of characters, instead of words and proved that such representations reveal information about the irony concealed in tweets. \par
In this work, we propose the combination of word- and character-level representations in order to exploit both semantic and syntactic information of each tweet for successfully predicting irony. For this purpose, we employ a deep LSTM architecture which models words and characters separately. We predict whether a tweet is ironic or not, as well as the type of irony in the ironic ones by ensembling the two separate models (late fusion). Furthermore, we add an attention layer to both models, to better weigh the contribution of each word and character towards irony prediction, as well as better interpret the descriptive power of our models. Attention weighting also better addresses the problem of supervising learning on deep learning architectures. The suggested model was trained only on constrained data, meaning that we did not utilize any external dataset for further tuning of the network weights. \par 
The two deep-learning models submitted to SemEval-2018 Task 3 \enquote{Irony detection in English tweets} \cite{VanHee2016:SemEval} are described in this paper with the following structure: in Section \ref{sec:over} an overview of the proposed models is presented, in Section \ref{models-description} the models for tracking irony are depicted in detail, in Section \ref{expe-res} the experimental setup alongside with the respective results are demonstrated and finally, in Section \ref{concl} we discuss the performance of the proposed models. \par


\begin{figure}[t]
	\captionsetup{farskip=0pt} 
	\centerline{\includegraphics[width=1.0\columnwidth]{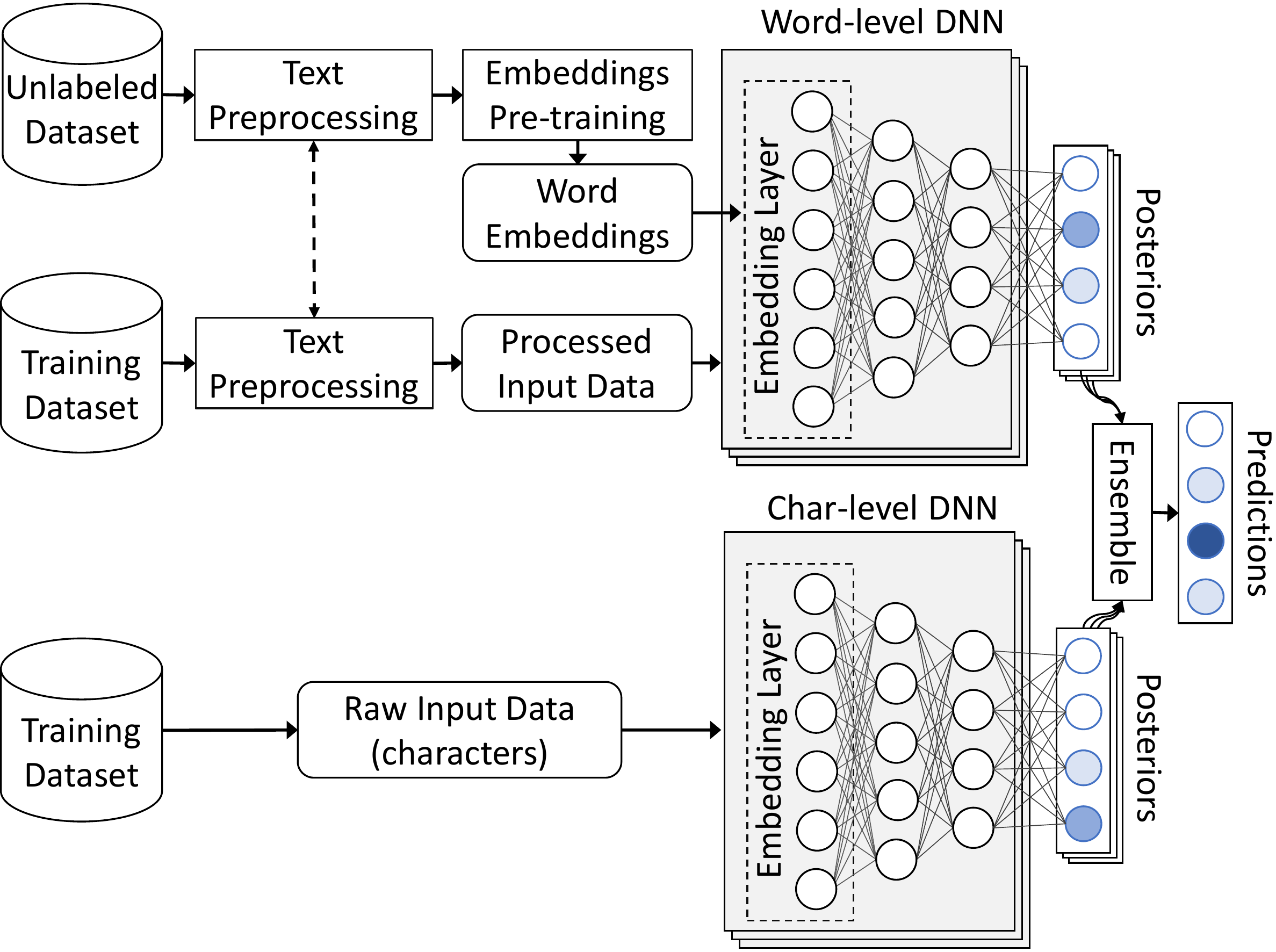}}
	\caption{High-level overview of our approach}
	\label{fig:over}
\end{figure}

\section{Overview}\label{sec:over}

Fig.~\ref{fig:over} provides a high-level overview of our approach, which consists of three main steps: 
\begin{enumerate*}[(1)]
	\item the \emph{pre-training of word embeddings}, where we train our own word embeddings on a big collection of unlabeled Twitter messages,
	\item the independent \emph{training of our models}: word- and char-level,
	\item the \emph{ensembling}, where we combine the predictions of each model.
\end{enumerate*}

\subsection{Task definitions}
The goal of Subtask A is tracking irony in tweets as a binary classification problem (ironic vs. non-ironic). In Subtask B, we are also called to determine the type of irony, with three different classes of irony on top of the non-ironic one (four-class classification). The types of irony are: \begin{enumerate*}[(1)]
	\item \textbf{Verbal irony by means of a polarity contrast}, which includes messages whose polarity (positive, negative) is inverted between the literal and the intended evaluation, such as  \textit{"I really love this year's summer; weeks and weeks of awful weather"},
where the literal evaluation (\textit{"I really love this year's summer"}) is positive, while the intended one, which is implied in the context (\textit{"weeks and weeks of awful weather"}), is negative.
	\item \textbf{Other verbal irony}, which refers to instances showing no polarity contrast, but are ironic such as \textit{"Yeah keeping cricket clean, that's what he wants \#Sarcasm"}
	 and \item \textbf{situational irony} which is present in messages that a present situation fails to meet some expectations, such as \textit{"Event technology session is having Internet problems. \#irony \#HSC2024"} in which the expectation that a technology session should provide Internet connection is not met.
\end{enumerate*}
 
\subsection{Data}
\noindent\textbf{Unlabeled Dataset}. We collected a dataset of 550 million archived English Twitter messages, from Apr. 2014 to Jun. 2017. This dataset is used for (1) calculating word statistics needed in our text preprocessing pipeline (Section \ref{sec:prep}) and (2) training word2vec word embeddings (Section \ref{sec:embeddings}).

\subsection{Word Embeddings} \label{sec:embeddings}
Word embeddings are dense vector representations of words~\cite{collobert2008, mikolov2013}, capturing semantic their and syntactic information.
We leverage our unlabeled dataset to train Twitter-specific word embeddings. 
We use the \textit{word2vec} ~\cite{mikolov2013} algorithm, with the skip-gram model, negative sampling of 5 and minimum word count of 20, utilizing Gensim's~\cite{rehurek_lrec} implementation. The resulting vocabulary contains $800,000$ words. 
The pre-trained word embeddings are used for initializing the first layer (embedding layer) of our neural networks.

{\setlength\extrarowheight{0.2em}
	\begin{table*}[!hb]
		\captionsetup{farskip=0pt} 
		\small
		\begin{tabularx}{\linewidth}{ |c|X| }
			\hline
			original  & The *new* season of \#TwinPeaks is coming on May 21, 2017. CANT WAIT \textbackslash o/ !!! \#tvseries \#davidlynch :D                                                                                         \\ 
			\hline
			processed & the new <emphasis> season of <hashtag> twin peaks </hashtag> is coming on <date> . cant <allcaps> wait <allcaps> <happy> ! <repeated> <hashtag> tv series </hashtag> <hashtag> david lynch </hashtag> <laugh> 
			\\ 
			\hline
		\end{tabularx}
		\caption{Example of our text processor}
		\label{table:textpp} 
	\end{table*}
}

\subsection{Preprocessing\footnote{Significant portions of the systems submitted to SemEval 2018 in Tasks 1, 2 and 3, by the NTUA-SLP team are shared, specifically the preprocessing and portions of the DNN architecture. Their description is repeated here for completeness.}}\label{sec:prep}
We utilized the \textit{ekphrasis}\footnote{\url{github.com/cbaziotis/ekphrasis}}~\cite{baziotis2017datastories} tool as a tweet preprocessor. The preprocessing steps included in ekphrasis are: Twitter-specific tokenization, spell correction, word normalization, word segmentation (for splitting hashtags) and word annotation.

\noindent\textbf{Tokenization}. 
Tokenization is the first fundamental preprocessing step and since it is the basis for the other steps, it immediately affects the quality of the features learned by the network. Tokenization in Twitter is especially challenging, since there is large variation in the vocabulary and the used expressions. Part of the challenge is also the decision of whether to process an entire expression (e.g. anti-american) or its respective tokens. 
Ekphrasis overcomes this challenge by recognizing the Twitter markup, emoticons, emojis, expressions like dates (e.g. 07/11/2011, April 23rd), times (e.g. 4:30pm, 11:00 am), currencies (e.g. \$10, 25mil, 50\euro), acronyms, censored words (e.g. s**t) and words with emphasis (e.g. *very*).

\noindent\textbf{Normalization}. After the tokenization we apply a series of modifications on the extracted tokens, such as spell correction, word normalization and segmentation. We also decide which tokens to omit, normalize and surround or replace with special tags (e.g. URLs, emails and @user).
For the tasks of spell correction \cite{jurafsky2000} and word segmentation \cite{segaran2009a} we use the Viterbi algorithm. The prior probabilities are initialized using uni/bi-gram word statistics from the unlabeled dataset. 

The benefits of the above procedure are the reduction of the vocabulary size, without removing any words, and the preservation of information that is usually lost during tokenization.
Table~\ref{table:textpp} shows an example text snippet and the resulting preprocessed tokens.

\subsection{Recurrent Neural Networks}
We model the Twitter messages using Recurrent Neural Networks (RNN). RNNs process their inputs sequentially, performing the same operation, $ h_t=f_W(x_t, h_{t-1}) $, on every element in a sequence,
where $h_t$ is the hidden state $t$ the time step, and $W$ the network weights. We can see that hidden state at each time step depends on previous hidden states, thus the order of elements (words) is important. This process also enables RNNs to handle inputs of variable length. 

RNNs are difficult to train \cite{pascanu2013a}, because gradients may grow or decay exponentially over long sequences \cite{bengio1994,hochreiter2001}. 
A way to overcome these problems is to use more sophisticated variants of regular RNNs, like Long Short-Term Memory (LSTM) networks~\cite{hochreiter1997} or Gated Recurrent Units (GRU)~\cite{cho2014a}, which introduce a gating mechanism to ensure proper gradient flow through the network. In this work, we use LSTMs.

\subsection{Self-Attention Mechanism}\label{sec:self-att}
\begin{figure}[!t]
	\captionsetup{farskip=0pt} 
	\centering
	\subfloat[Regular RNN ]{\includegraphics[width=0.48\columnwidth,page=1]{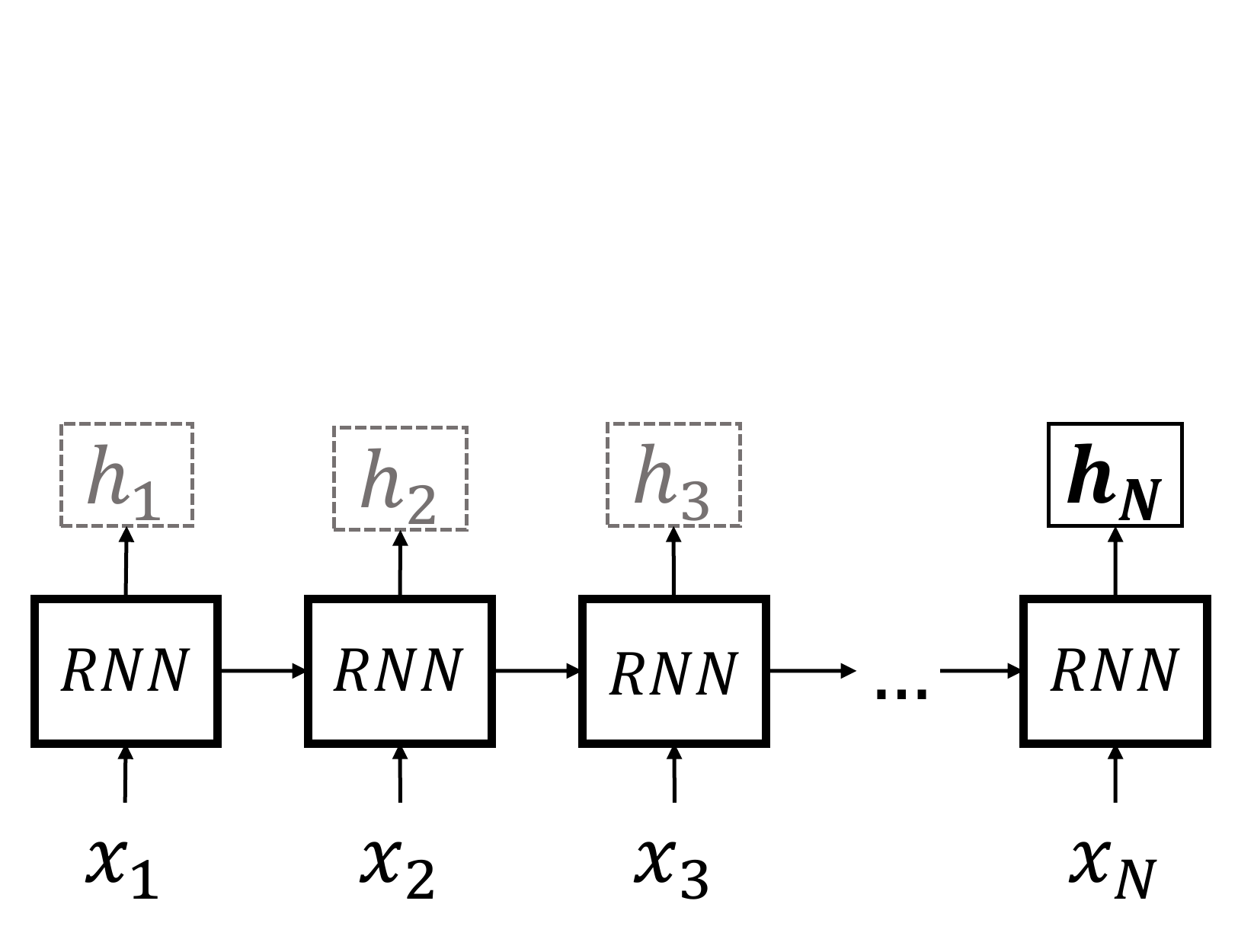}\label{fig:rnn}}
	\hfill
	\subfloat[Attention RNN]{\includegraphics[width=0.48\columnwidth,page=2]{rnn}\label{fig:rnn_att}}
	\caption{Comparison between the regular RNN and the RNN with attention.}
	\label{fig:attention}
\end{figure}

RNNs update their hidden state $h_i$ as they process a sequence and the final hidden state holds a summary of the information in the sequence. 
In order to amplify the contribution of important words in the final representation, a self-attention mechanism \cite{DBLP:journals/corr/BahdanauCB14} can be used (Fig.~\ref{fig:attention}). 
In normal RNNs, we use as representation $r$ of the input sequence its final state $h_N$. However, using an attention mechanism, we compute $r$ as the convex combination of all $h_i$. The weights $a_i$ are learned by the network and their magnitude signifies the importance of each hidden state in the final representation. Formally:
$
r = \sum_{i=1}^{N} a_i h_i,  \mbox{where} \sum_{i=1}^{N} a_i = 1, \; \mbox{and} \; a_i > 0.
$

\section{Models Description}\label{models-description}
We have designed two independent deep-learning models, with each one capturing different aspects of the tweet. The first model operates at the word-level, capturing the semantic information of the tweet and the second model at the character-level, capturing the syntactic information. Both models share the same architecture, and the only difference is in their embedding layers. We present both models in a unified manner.

\subsection{Embedding Layer}\label{sec:char_model}

\noindent\textbf{Character-level}.
The input to the network is a Twitter message, treated as a sequence of characters. We use a character embedding layer to project the characters $c_1,c_2,...,c_N$ to a low-dimensional vector space $ R^C$, where $C$ the size of the embedding layer and $N$ the number of characters in a tweet. We randomly initialize the weights of the embedding layer and learn the character embeddings from scratch.

\noindent\textbf{Word-level}.
The input to the network is a Twitter message, treated as a sequence of words. We use a word embedding layer to project the words $w_1,w_2,...,w_N$ to a low-dimensional vector space $ R^W$, where $W$ the size of the embedding layer and $N$ the number of words in a tweet. We initialize the weights of the embedding layer with our pre-trained word embeddings.

\subsection{BiLSTM Layers}
An LSTM takes as input the words (characters) of a tweet and produces the word (character) annotations $h_1,h_2,...,h_N$, where $h_i $ is the hidden state of the LSTM at time-step $i$, summarizing  all the information of the sentence up to $w_i$ ($c_i$). 
We use bidirectional LSTM (BiLSTM) in order to get word (character) annotations that summarize the information from both directions. A bidirectional LSTM consists of a forward LSTM $ \overrightarrow{f} $  that reads the sentence from $w_1$ to $w_N$ and a backward LSTM $ \overleftarrow{f} $ that reads the sentence from $w_N$ to $w_1$. We obtain the final annotation for a given word $w_i$ (character $c_i$), by concatenating the annotations from both directions,
$
h_i = \overrightarrow{h_i} \parallel \overleftarrow{h_i}, \quad h_i \in R^{2L}
$
where $ \parallel $ denotes the concatenation operation and $L$ the size of each LSTM. 
We stack two layers of BiLSTMs in order to learn more high-level (abstract) features. 

\subsection{Attention Layer}
Not all words contribute equally to the meaning that is expressed in a message. We use an attention mechanism  to find the relative contribution (importance) of each word. The attention mechanism assigns a weight $a_i$ to each word annotation $h_i$. We compute the fixed representation $r$ of the whole input message. as the weighted sum of all the word annotations.
\begin{align}
e_i &= tanh(W_h h_i + b_h)\label{eq:att_ei}, \quad e_i \in [-1,1]\\
a_i &= \dfrac{exp(e_i)}{\sum_{t=1}^{T} exp(e_t)}\label{eq:att_ai}, \quad \sum_{i=1}^{T} a_i = 1\\
r &= \sum_{i=1}^{T} a_ih_i \label{eq:att_r}, \quad r \in R^{2L}
\end{align}
where $ W_h $ and $ b_h $ are the attention layer's weights.

\noindent\textbf{Character-level Interpretation}. In the case of the character-level model, the attention mechanism operates in the same way as in the word-level model. However, we can interpret the weight given to each character annotation $h_i$ by the attention mechanism, as the importance of the information surrounding the given character.

\subsection{Output Layer}
We use the representation $r$ as feature vector for classification and we feed it to a fully-connected softmax layer with $L$ neurons, which outputs a probability distribution over all classes $p_c$ as described in Eq.~\ref{e:outlay}:

\begin{equation}
\label{e:outlay}
p_c = \frac{e^{Wr + b}}{\sum_{i \in [1,L]}(e^{W_i r + b_i})}
\end{equation}
where $W$ and $b$ are the layer's weights and biases.

\begin{figure}[tbh]
	\centering
	\includegraphics[trim={5pt 5pt 5pt 10pt},clip,width=0.95\columnwidth, page=2]{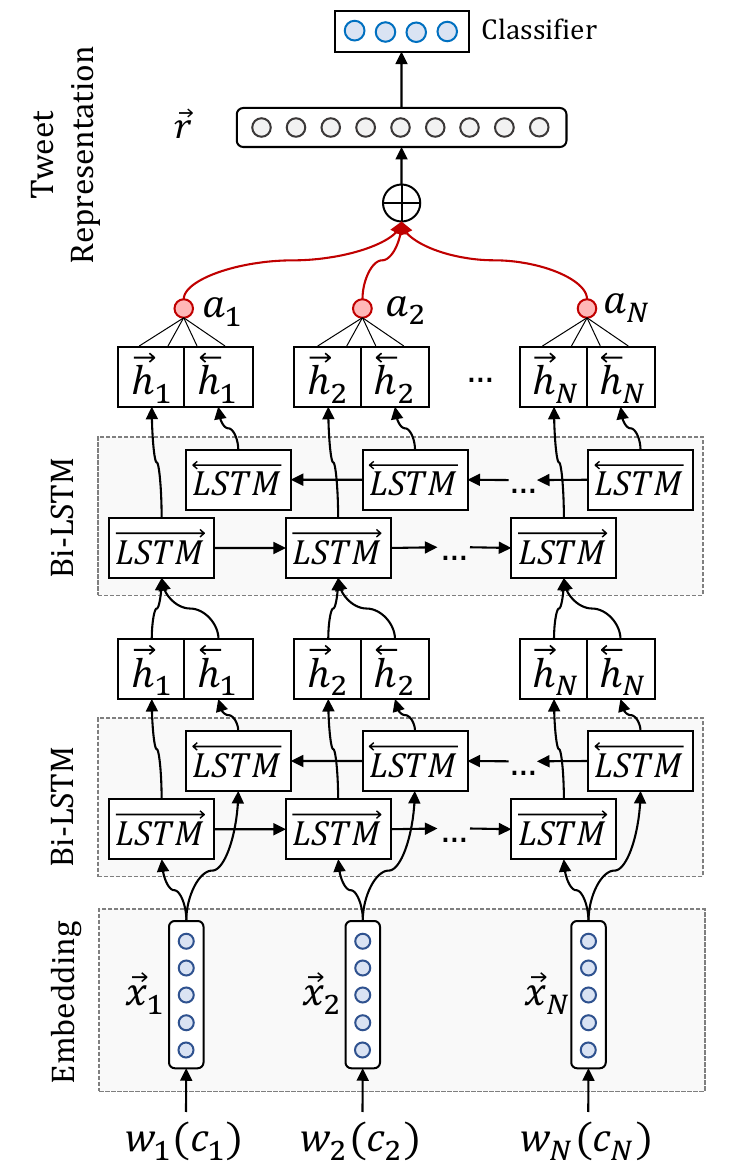}
	\caption{The word/character-level model.}
	\label{fig:word_model}
\end{figure}

\subsection{Regularization}\label{sec:reg}
In order to prevent overfitting of both models, we add Gaussian noise to the embedding layer, which can be interpreted as a random data augmentation technique, that makes models more robust to overfitting.
In addition to that, we use dropout \cite{srivastava2014} and early-stopping.

Finally, we do not fine-tune the embedding layers of the word-level model. Words occurring in the training set, will be moved in the embedding space and the classifier will correlate certain regions (in embedding space) to certain meanings or types of irony. However, words in the test set and not in the training set, will remain at their initial position which may no longer reflect their ``true'' meaning, leading to miss-classifications.

\subsection{Ensemble} \label{sec::ensemble}
A key factor to good ensembles, is to utilize diverse classifiers. To this end, we combine the predictions of our word and character level models. We employed two ensemble schemes, namely unweighted average and majority voting.

\noindent\textbf{Unweighted Average (UA)}.
In this approach, the final prediction is estimated from the unweighted average of the posterior probabilities for all different models.  Formally, the final prediction $p$ for a training instance is estimated by:
\begin{align}
p &= \argmax_{c}\dfrac{1}{C} \sum_{i=1}^{M} \vec{p_i} \space ,\quad p_i \in {\rm I\!R}^C \label{eq:average}
\end{align}
where $C$ is the number of classes, $M$ is the number of different models, $c \in \{1,...,C\}$ denotes one class and $\vec{p_i}$ is the probability vector calculated by model $i \in \{1,...,M\}$ using softmax function. \par
\noindent\textbf{Majority Voting (MV)}.
Majority voting approach counts the votes of all different models and chooses the class with most votes. Compared to unweighted averaging, MV is affected less by single-network decisions. However, this schema does not consider any information derived from the minority models. Formally, for a task with $C$ classes and $M$ different models, the prediction for a specific instance is estimated as follows: 
\begin{align}
\label{eqn:eqlabel}
\begin{split}
v_c &=  \sum_{i=1}^{M} F_i(c) \\
p &= \argmax_{c \in \{1,...,C\}} v_c 
\end{split}
\end{align}
where $v_c$ denotes the votes for class $c$ from all different models, $F_i$ is the decision of the $i^{th}$ model, which is either 1 or 0 with respect to whether the model has classified the instance in class $c$ or not, respectively, and $p$ is the final prediction.

\section{Experiments and Results} \label{expe-res}

\subsection{Experimental Setup}

\noindent\textbf{Class Weights}. 
In order to deal with the problem of class imbalances in Subtask B, we apply
class weights to the loss function of our models, penalizing more the misclassification of underrepresented
classes. We weight each class by its inverse frequency in the training set.

\noindent\textbf{Training}\label{sec:train}
We use Adam algorithm~\cite{kingma2014} for optimizing our networks, with mini-batches of size 32 and we clip the norm of the gradients~\cite{pascanu2013a} at 1, as an extra safety measure against exploding gradients. For developing our models we used PyTorch \cite{paszke2017automatic} and Scikit-learn \cite{pedregosa2011}.

\noindent\textbf{Hyper-parameters}.
In order to find good hyper-parameter values in a relative short time (compared to grid or random search), we adopt the Bayesian optimization \cite{bergstra2013} approach, performing a ``smart'' search in the high dimensional space of all the possible values. 
Table \ref{hparams}, shows the selected hyper-parameters.

\begin{table}[!htb]

	\captionsetup{farskip=0pt} 
    \centering
    \small
    \begin{tabular}{|l|c|c|}
    \hline
                          & \multicolumn{1}{l|}{\textbf{Word-Model}} & \multicolumn{1}{l|}{\textbf{Char-Model}} \\ \hline
    \textbf{Embeddings}   & 300                                      & 25                                       \\ \hline
    \textbf{Emb. Dropout} & 0.1                                      & 0.0                                      \\ \hline
    \textbf{Emb. Noise}   & 0.05                                     & 0.0                                      \\ \hline
    \textbf{LSTM (x2)}    & 150                                      & 150                                      \\ \hline
    \textbf{LSTM Dropout} & 0.2                                      & 0.2                                      \\ \hline
    \end{tabular}
    
    \caption{Hyper-parameters of our models.}\label{hparams}
\end{table}

\subsection{Results and Discussion}
Our official ranking is 2/43 in Subtask A and 2/29 in Subtask B as shown in Tables \ref{tab:resA} and \ref{tab:resB}. Based on these rankings, the performance of the suggested model is competitive on both the binary and the multi-class classification problem. Except for its overall good performance, it also presents a stable behavior when moving from two to four classes.  

\begin{table}[!h]
\captionsetup{farskip=0pt} 
\small
\centering
\begin{tabular}{|p{0.2em}|p{1.9cm}|p{0.75cm}|p{0.75cm}|p{0.75cm}|p{0.75cm}|}
\hline 
\multicolumn{1}{|l|}{\textbf{\#}} & \textbf{Team Name} & \multicolumn{1}{c|}{\textbf{Acc}} & \multicolumn{1}{c|}{\textbf{Prec}} & \multicolumn{1}{c|}{\textbf{Rec}} & \multicolumn{1}{c|}{\textbf{F1}} \\ \hline
1 & THU\_NGN & 0.7347 & 0.6304 & 0.8006 & 0.7054 \\ \hline
2 & \textbf{NTUA-SLP} & 0.7321 & 0.6535 & 0.6913 & 0.6719 \\ \hline
3 & WLV & 0.6429 & 0.5317 & 0.8360 & 0.6500 \\ \hline
4 & \textit{Unknown} & 0.6607 & 0.5506 & 0.7878 & 0.6481 \\ \hline
5 & NIHRIO, NCL & 0.7015 & 0.6091 & 0.6913 & 0.6476 \\ \hline
\end{tabular}
\caption{Competition results for Subtask A}
\label{tab:resA}
\end{table}

\begin{table}[!h]
	\small
	\centering
	\begin{tabular}{|p{0.2em}|p{1.9cm}|p{0.75cm}|p{0.75cm}|p{0.75cm}|p{0.75cm}|}
		\hline 
		\multicolumn{1}{|l|}{\textbf{\#}} & \textbf{Team Name} & \multicolumn{1}{c|}{\textbf{Acc}} & \multicolumn{1}{c|}{\textbf{Prec}} & \multicolumn{1}{c|}{\textbf{Rec}} & \multicolumn{1}{c|}{\textbf{F1}} \\ \hline
		\multicolumn{1}{|c|}{1}           & \textit{Unknown}   & 0.7321                            & 0.5768                             & 0.5044                            & 0.5074                           \\ \hline
		\multicolumn{1}{|c|}{2}           & \textbf{NTUA-SLP}  & 0.6518                            & 0.4959                             & 0.5124                            & 0.4959                           \\ \hline
		\multicolumn{1}{|c|}{3}           & THU\_NGN           & 0.6046                            & 0.4860                             & 0.5414                            & 0.4947                           \\ \hline
		\multicolumn{1}{|c|}{4}           & \textit{Unknown}   & 0.6033                            & 0.4660                             & 0.5058                            & 0.4743                           \\ \hline
		\multicolumn{1}{|c|}{5}           & NIHRIO, NCL        & 0.6594                            & 0.5446                             & 0.4475                            & 0.4437                           \\ \hline
	\end{tabular}
	\caption{Competition results for Subtask B}
	\label{tab:resB}
\end{table}

\begin{table}[!h]
	\small
	\centering
	\begin{tabular}{|p{1.6cm}|c|c|c|c|}
		\hline
		\textbf{model}  & \textbf{Acc}    & \textbf{Prec}   & \textbf{Rec}    & \textbf{f1}     \\
		\hline
		\hline
		BOW             & 0.6531          & 0.6453          & 0.6417          & 0.6426          \\
		\hline
		N-BOW           & 0.6645          & 0.6543          & 0.6517          & 0.6527          \\
		\hline
		\hline
		LSTM-char       & 0.6241          & 0.6371          & 0.6342          & 0.6163          \\
		\hline
		LSTM-word       & 0.7746          & 0.7726          & 0.7826          & 0.7698          \\
		\hline
		\hline
		\textbf{Ens-MV} & 0.7462          & 0.7381          & 0.7461          & 0.7400          \\
		\hline
		\textbf{Ens-UA} & \textbf{0.7883} & \textbf{0.7865} & \textbf{0.7992} & \textbf{0.7856} \\
		\hline
	\end{tabular}
	\caption{Results of our models for Subtask A}
	\label{table:results_a}
\end{table}


\begin{table}[!h]
	\small
	\centering
	\begin{tabular}{|p{1.6cm}|c|c|c|c|}
		\hline
		\textbf{model}  & \textbf{Acc}    & \textbf{Prec}   & \textbf{Rec}    & \textbf{f1}     \\
		\hline
		\hline
		BOW             & 0.5880          & 0.4460          & 0.4384          & 0.4371          \\
		\hline
		N-BOW           & 0.6084          & 0.4649          & 0.4560          & 0.4520          \\
		\hline
		\hline
		LSTM-char       & 0.5726          & 0.4098          & 0.4102          & 0.3782          \\
		\hline
		LSTM-word       & \textbf{0.6987} & 0.5394          & \textbf{0.5790} & 0.5315          \\
		\hline
		\hline
		\textbf{Ens-MV} & 0.6888          & \textbf{0.5433} & 0.5442          & \textbf{0.5358} \\
		\hline
		\textbf{Ens-UA} & 0.6888          & 0.5361          & 0.4874          & 0.4959          \\
		\hline
	\end{tabular}
	\caption{Results of our models for Subtask B}
	\label{table:results_b}
\end{table}


Additional experimentation following the official submission significantly improved the efficiency of our models. The results of this experimentation, tested on the same data set, are shown in Tables \ref{table:results_a} and \ref{table:results_b}. The first baseline is a Bag of Words (BOW) model with TF-IDF weighting. The second baseline is a Neural Bag of Words (N-BOW) model where we retrieve the word2vec representations of the words in a tweet and compute the tweet representation as the centroid of the constituent word2vec representations.
Both BOW and N-BOW features are then fed to a linear SVM classifier, with tuned $C=0.6$. 

The best performance that we achieve, as shown in Tables \ref{table:results_a} and \ref{table:results_b}  is \textbf{0.7856} and \textbf{0.5358} for Subtask A and B respectively\footnote{The reported performance is boosted in comparison with the results presented in Tables \ref{tab:resA} and \ref{tab:resB} due to the utilization of unnormalized word vectors. Specifically, after further experimentation we found that normalization of word vectors provided to the LSTM is detrimental to performance, because semantic information is encoded by both the angle and length of the embedding vectors~\cite{wilson2015controlled}.}\footnote{For our DNNs, the results are computed by averaging $10$ runs to account for the variability in training performance.}. 
In Subtask A the BOW and N-BOW models perform similarly with respect to f1 metric and word-level LSTM is the most competitive individual model. However, the best performance is achieved when the character- and the word-level LSTM models are combined via the unweighted average ensembling method, showing that the two suggested models indeed contain different types of information related to irony on tweets. Similar observations are derived for Subtask B, except that the character-level model in this case performs worse than the baseline models and contributes less to the final results.

\begin{figure*}[!t]
	\captionsetup{farskip=0pt} 
	\centering

	\subfloat{\includegraphics[scale=0.7,page=22]{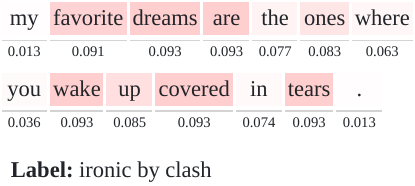}\label{fig:type_irony_1}}
	\hfill
	\subfloat{\includegraphics[scale=0.7,page=62]{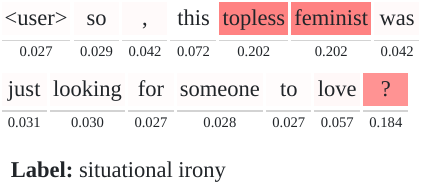}\label{fig:type_irony_2}}
	\hfill
	\subfloat{\includegraphics[scale=0.7,page=19]{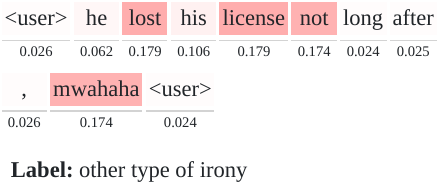}\label{fig:type_irony_3}}

	\caption{Examples of the attention mechanism for identification of the type of irony in each sentence.}
	\label{fig:irony_types}
\end{figure*}

\begin{figure*}[!t]

    \centering
    
	\begin{minipage}{.45\textwidth}
      	\hspace{50pt}\includegraphics[scale=0.45,page=79]{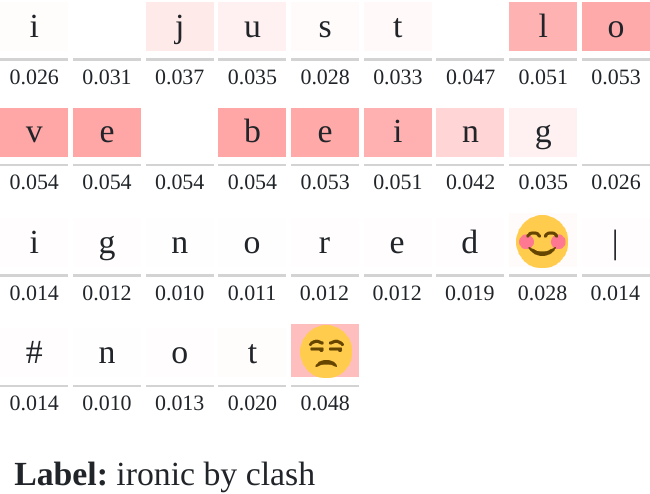}
        \label{fig:irony_word_1}
        \bigbreak
        \hspace{50pt}\includegraphics[scale=0.7,page=16]{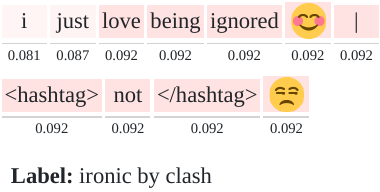}
        \label{fig:irony_char_1}
	\end{minipage}%
    \hfill
	\begin{minipage}{0.45\textwidth}
      	\hspace{30pt}\includegraphics[scale=0.45,page=93]{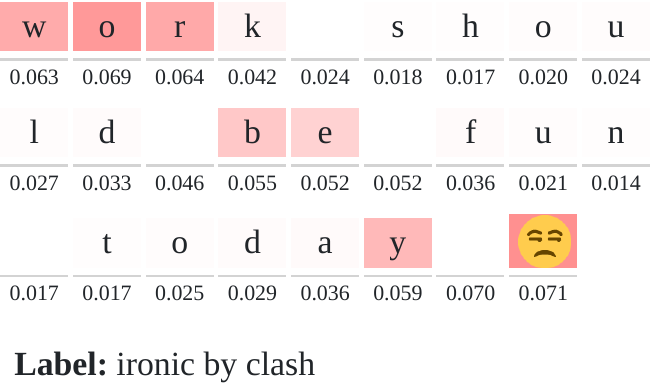}
        \label{fig:irony_word_2}
        \bigbreak
        \bigbreak
        \vspace{5pt}
        \hspace{30pt}\includegraphics[scale=0.45,page=143]{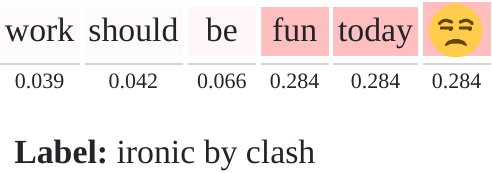}
        \bigbreak
        \bigbreak
        \vspace{12pt}
        \label{fig:irony_char_2}
	\end{minipage}
    
\caption{Comparison of the behavior of the word and character level models.}\label{fig:word_char_att}

\end{figure*}

\subsection{Attention Visualizations}
Our models' behavior can be interpreted by visualizing the distribution of the attention weights assigned to the words (characters) of the tweet. The weights signify the contribution of each word (character), to model's final classification decision.
In Fig.~\ref{fig:irony_types}, examples of the weights assigned by the word level model to ironic tweets are presented. The salient keywords that capture the essence of irony or even polarity transitions (e.g. irony by clash) are correctly identified by the model. 
Moreover, in Fig.~\ref{fig:word_char_att} we compare the behavior of the word and character models on the same tweets. In the first example, the character level model assigns larger weights to the most discriminative words whereas the weights assigned by the word level model seem uniform and insufficient in spotting the polarity transition. However, in the second example, the character level model does not attribute any weight to the words with positive polarity (e.g. ``fun'') compared to the word level model. Based on these observations, the two models indeed behave diversely and consequently contribute to the final outcome (see Section \ref{sec::ensemble}).
\section{Conclusion}\label{concl}
In this paper we present an ensemble of two different deep learning models: a word- and a character-level deep LSTM for capturing the semantic and syntactic information of tweets, respectively. We demonstrated that combining the predictions of the two models yields competitive results in both subtasks for irony prediction. Moreover, we proved that both types of information (semantic and syntactic) contribute to the final results with the word-level model, however, individually achieving more accurate irony prediction. Also, the best way of combining the outcomes of the separate models is by conducting majority voting over the respective posteriors. Finally, the proposed model successfully predicts the irony in tweets without exploiting any external information derived from hand-crafted features or lexicons. \par
The performance reported in this paper could be further boosted by utilizing transfer learning methods from larger datasets. Moreover, the joint training of word- and character-level models can be tested for further improvement of the results.  
Finally, we make the source code of our models and our pretrained word embeddings available to the community\footnote{\url{github.com/cbaziotis/ntua-slp-semeval2018-task3}}, 
in order to make our results easily reproducible and facilitate further experimentation.
\newline

\noindent\textbf{Acknowledgements}. 
This work has been partially supported by the BabyRobot project supported by EU H2020 (grant \#687831). Also, the authors would like to thank NVIDIA for supporting this work by donating a TitanX GPU.

\bibliography{refs}
\bibliographystyle{acl_natbib}

\appendix

\end{document}